# Annotate Rhetorical Relations with INCEpTION: A Comparison with Automatic Approaches

Mehedi Hasan Emon
Linguistic Data Science Lab
Ruhr University Bochum
Email: Mehedi.Emon@ruhr-uni-bochum.de

## Abstract

Automatically identifying rhetorical relations in discourse units is a challenging task in natural language processing (NLP) because it should be able to logically and semantically connect the discourse units. Although large language models (LLMs) shows potential for application in many domains, including text classification tasks, their effectiveness in predicting rhetorical relations remains open for research. One of the major challenges in this domain is the lack of annotated data sets capturing different rhetorical relations, which would then make model training more difficult.

In this research, we manually created the datasets from various cricket reports and then annotated the reports as discourse units. We used the INCEpTION annotation tools for annotation and then structured the dataset for the machine-learning model. BERT and DistilBERT were then used to classify rhetorical relations, and their performance was evaluated based on accuracy, and F1-score. Additionally, a Logistic Regression model was used as a baseline model.

The results suggest that DistilBERT provide the highest accuracy, while BERT struggle to classify some relationships which we discover with the Error analysis. It is often confused with the cause-effect relationship, while DistilBERT is not particularly confused with any relationship.

This study highlights the potential of large language models for predicting rhetorical relations while pointing out the need for larger datasets and suggesting that domain-based fine-tuning would be necessary for further performance improvement.

## 1. Introduction

Discourse parsing is a major task in natural language processing (NLP) that deals with the analysis of text to identify its structure and the relations between these components. Since human communication naturally relies on the interaction of rhetorical relations to convey meaning, the ability to parse these relations accurately has great implications for a wide variety of downstream applications, such as text summarization, sentiment analysis, machine translation, and question-answering systems. Still, even with the progress within NLP, annotation and automated identification of rhetorical relations, given the balance between human interpretability and machine learning efficiency, remain challenging.

Here is an example of discourse Parsing from our study:-

"England pace bowler Jofra Archer could play in this year's T20 World Cup but will not play test cricket until 2025, according to England managing director Rob Key." (The Daily Star, Dhaka)

**EDU1**: England pace bowler Jofra Archer could play in this year's T20 World Cup but

**EDU2**: will not play test cricket until 2025,

**EDU3**: according to England managing director Rob Key.

Here, we split our sentence into the smallest unit which can link different parts of the text. EDUs refers to "elementary discourse units" which is mention by paper "Rhetorical Structure Theory" by Mann and Thompson (1988) and Taboada and Mann (2006). They propose that a coherent text can be represented using a tree structure, where the leaves are the EDUs and the internal nodes are labeled with coherence relations.

The annotation of the example EDUs is as follows:

"Background(Contrast(EDU1, EDU2),EDU3)"

From our study, we conclude that EDU1 and EDU2 have a contrast relationship, while EDU3 has a background relationship.

## 1.1 Importance of Rhetorical Relations in Discourse Parsing

Rhetorical relations refer to the logical and semantic connections between the different parts of a discourse. These connections are the basis for coherent communication, whereby the innermost meaning of complex texts can be understood by the reader and listener themselves. Take "elaboration" into account, where one item is explained with respect to another; "contrast," which identifies opposite ideas; and "cause-effect," which denotes causation between events.

Understanding these types of relations is very important for the interpretation of the text's deeper meanings. For example, in a sports report, the call of "background" would indicate the setup with which a match takes place, while a complete identification of "contrast" relations should direct most important differentiators between competing teams. When a system has been able to correctly parse discourse, it becomes capable of achieving human-like understanding of text, making it possible to summarize the sports report or generate coherent responses in dialogue systems.

## 1.2 Research Gap and Significance

Discourse parsing has many frameworks and tools dedicated to it, for example, the Rhetorical Structure Theory by Mann and Thompson (1988) and the Penn Discourse Treebank by Miltsakaki and Prasad (2004). Yet, challenges remain in the field. For example, RST uses the structural paradigm of tree structures to show the various connections defined between elementary discourse units (EDUs). The problem related to this type of relation is that the annotation process to be executed manually is labor intensive as well as subjective. Now a days we have more advance annotation tools such as "INCEpTION" by Klie and Bugert (2018). Most existing studies focus on either manual annotation or automated processes, with little comparison made between the two methodologies, especially when using annotation tools like INCEpTION.

Introducing large language models like BERT and DistilBERT revolutionized many NLP tasks including predicting rhetorical relationship from discourse parsing. Models trained on gigantic corpora have thus proven their worth as miraculous semantic understanding of context. However, little has been done in examining their use cases for rhetorical relations over-and-above the human benchmark. This study closes the gap by contrasting human with LLM-based predilections for a selected number of rhetorical relations such as "elaboration", "background", "contrast", "narration", "concession", "restatement", "cause-effec", and "joint".

## 2. Review Related works

The study of discourse parsing and understand rhetorical relations from discourse parsing has been explored by various scholars, The interest in the research of rhetorical relations has been one of the major concerns of discourse analysis and natural language processing. One of the pioneering works in this area is that of Mann and Thompson (1988), who are credited with having founded Rhetorical Structure Theory (RST). His work offers a very good frame in analyzing the structure of texts in terms of rhetorical linking between elementary discourse units (EDUs). The Mann and Thompson approach refers to the hierarchical organization of discourse where relations such as "elaboration", "contrast", "cause-effect", etc as contribute to the understandability and coherence of text. This research remains a central pillar in the field and becomes the theoretical base for much of the further research in this area of discourse parsing.

Another contribution is from Hu and Wan (2023): RST Discourse Parsing as Text-to-Text Generation. This study analyzes using large language models (LLMs) in discourse parsing by formulating it as a task of converting one text into another. The approach takes advantage of transformer architectures like T5 to show that LLMs can effectively generate discourse structure and identify rhetorical relations. The work represents a break from conventional parsing, which is done using feature engineering and statistical models. Rather Lu and Wan's approach capitalizes on the contextual understanding and generative capabilities of LLMs for a more scalable and flexible solution to the problems of discourse parsing.

Moreover, Stede et al. (2017) have contributed to this field with Annotation Guidelines for Rhetorical Structure. Their contribution has a comprehensive framework for annotating rhetorical relations, with a main emphasis on consistency and reproducibility in annotation. Annotation guidelines have strongly advised segmenting texts into elementary discourse units (EDUs) with relation types such as "elaboration," "contrast," and "cause-effect." Besides the authors also discussed about the understanding of ambiguity in identifying relations, strategies for disambiguation and for maintaining annotation quality. This work has served as a weapon towards the establishment of standardized practices for manual discourse annotation, which has become an important reference source for rhetorical projects.

Then again, in their work Annotation Guidelines for Rhetorical Structure, Stede et al. (2017) did too much toward the discourse parsing field. For example, they created a well-structured annotation framework for annotating rhetorical relations with a primary emphasis on consistency and reproducibility in the annotation process. They recommended segmenting text into elementary discourse units (EDUs) and assigning to them relations such as "elaboration," "contrast," and "cause-effect." The authors further addressed the ambiguity problems of relation identification, offering strategies for disambiguation and maintaining quality of annotation. This valuable work has set up a standard procedure manual on discourse annotation and is targeted as an essential reference for projects focusing on rhetorical relations.

More specifically, Stede et al. (2017) contributed to this field by writing Annotation Guidelines for Rhetorical Structure. Their contribution had a thorough framework for annotating rhetorical relations with a major emphasis on annotating methods that are consistent and reproducible. The annotation guidelines strongly recommended that text should be segmented into elementary discourse units (EDUs) associated with relation types, such as elaboration, contrast, and cause-effect. The issues of ambiguity from relations were discussed very well by authors along with the strategies for disambiguation and maintaining quality in annotation. This work has been instrumental in developing the standard practices of manual discourse annotation and further serves as an important reference in projects dealing with rhetorical relations.

In addition, Stede et al. (2017) have greatly contributed to the discourse parsing field by publishing their work Annotation Guidelines for Rhetorical Structure. For example, they had established a robust framework for annotating rhetorical relations, with a main emphasis on annotation consistency and reproducibility. They have strongly recommended segmenting texts into elementary discourse units (EDUs) with relations such as "elaboration," "contrast," and "cause-effect." Besides the authors also discussed about the understanding of ambiguity in identifying relations, strategies for disambiguation, and for maintaining annotation quality. This work has served as a weapon towards establishing standardized practices for manual discourse annotations which have become an important reference source for rhetorical projects.

Klie et al. (2018) proposed the INCEpTION platform, which is a machine-assisted and knowledge-oriented interactive annotation tool that has undoubtedly changed the annotation process of an NLP task. It gives the user the capability to annotate texts interactively, which can then rely on machine learning models to help suggestion and prediction aspects. Therefore, it is a flexible platform that can fit any annotation scheme that makes it the best example in annotating rhetorical relations within complex datasets.

As mentioned in the study, INCEpTION made annotation efficient by offering features like active learning, real-time feeds, and adaptable workflows. Most of these functions benefit the segmentation of a text into elementary discourse units (EDUs) and labeling rhetorical connections as required in RST-based annotation tasks. With the application of INCEpTION for this work, annotating rhetorical relations in sports reports such as cricket report can be streamlined and made efficient while ensuring that such an exercise is consistent and accurate.

## 3. Methods

### 3.1 Solution Architecture

To conduct the RST study, the experiment was divided into multiple sub-tasks. The workflow followed a sequential process, where each task needed to be completed before proceeding to the next. Consequently, each sub-task was interdependent.

### 3.2 Environment Setup

The INCEpTION Annotation Tool (Klie et al., 2018) version 34.5 was utilized to annotate rhetorical relations within discourse units. Most of the implementation of AI models and data preprocessing was conducted using the Python 3.12.2 environment. To predict rhetorical relations from the discourse units, our most effective approaches involved two pre-trained large language models: BERT (Devlin et al., 2019) and DistilBERT (Sanh et al. 2019), along with a logistic regression model. These pre-trained models were accessed via the Hugging Face hub. Data processing and numerical operations were performed using Pandas 1.5.2 and NumPy 1.24.0. The training, validation, and evaluation of the models took place on a Google Colab runtime, which allowed for enhanced computational power and faster execution.

Additionally, part of the preprocessing and experimentation was conducted locally on a laptop equipped with an Intel(R) Core(TM) i5-8350U CPU running at 1.70 GHz (1.90 GHz boost) and 16 GB of RAM.

For visualizing the results from the various models, we used Matplotlib, specifically utilizing pyplot, box plots, subplots, and other visualization tools.

The dataset's for this experiment comprised a collection of 10 sports reports, annotated with discourse units and their corresponding relations. Further details will be provided in later sections.

### 3.3 Data collection

This research project is based on a sports dataset's focused on cricket news reports, which have been collected from several reputable news websites. Although the total number of sports reports for this experiment is not extensive as we created our own datasets, so we selected 10 reports that together created 57 discourse units. These reports were chosen primarily because of their systematic organization and the presence of a maximum number of rhetorical relations. The report have been divided into elementary discourse units (EDUs), which will serve as the basis for both manual and automated analysis.

#### 3.3.1 Data Preparation

The data preparation process was essential for organizing and cleaning the raw text so that it could be utilized effectively in both manual and automated discourse parsing. Initially, the sports reports stored in plain text files were processed to transformed into elementary discourse units (EDUs), accompanied by their respective rhetorical relations. The data was ultimately stored in structured formats, such as CSV files, to facilitate more efficient model training and evaluation.

Each CSV entry includes the following columns:

**CSV Dataset:**

| EDU1 | EDU2 | Label |
|---|---|---|
| England pace bowler Jofra Archer could play in this year's T20 World Cup but | will not play test cricket until 2025 | Contrast |
| England pace bowler Jofra Archer could play in this year's T20 World Cup but will not play test cricket until 2025 | according to England managing director Rob Key. | Background |

*Tabel 1: partial view of the dataset (Annotated)*

By preparing the data in this structured manner, the research ensures compatibility with machine learning models and simplifies the annotation process. This approach also provides a consistent framework for comparing manual and automated discourse parsing results.

#### 3.3.2 Overview of Dataset:

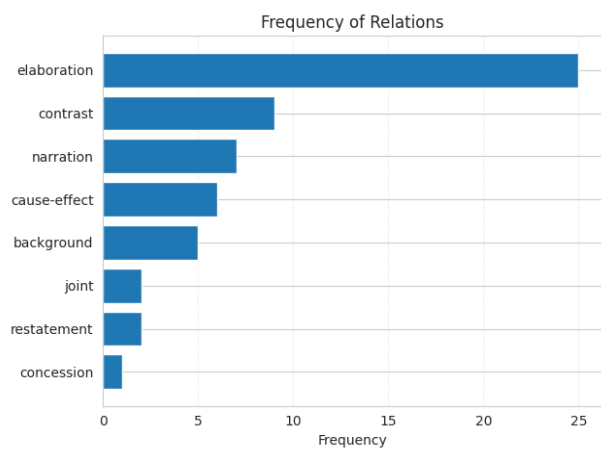

*Figure 1: Overview of the data with RST relations*

Figure 1, shows the overall overview of our sports data and its frequency of the different RST relations.

### 3.4 INCEpTION

The INCEpTION annotation tools used for this study are used to perform annotations between discourse units, an advanced annotation tool designed to facilitate discourse analysis and relation extraction tasks. IN-CEpTION provides a browser-based frontend, which is accessed locally via a local host server. This user-friendly interface allows annotators to efficiently define relationships between discourse units and manage annotations with administrative privileges.

### 3.4.1 Setup and Configuration

To prepare for annotation, administrative access in INCEpTION tools was utilized to configure the "Layers" required for the task. Two primary layers were defined Span & Relations. The span Layer is used to identify and label individual discourse units and it represents a segment of text (e.g., an EDU) that forms a meaningful part of the discourse structure. The relations Layer is used to establish rhetorical relationships between spans. Each relation includes a source (starting discourse unit), a target (related discourse unit), and a label representing the type of relationship (e.g., Elaboration, Contrast). The spans and relations layers were configured to store annotations as strings, enabling flexible and descriptive labeling of discourse elements.

**Annotation Process:**

The annotation activities were carried out through an interactive interface of the platform. Such activities comprise selecting discourse units from the discourse text and assigning them as spans. Identifying rhetorical relations between these spans and annotating them according to the predefined labels. Furthermore, refining annotations using the platform's machine learning capabilities. INCEpTION comes with a machine-learning approach that learns from the annotator's input to suggest the most probable relationships, thus rendering the process more robust and efficient.

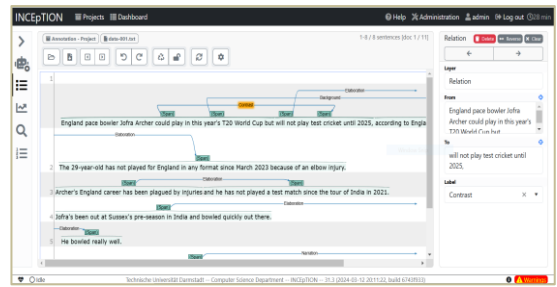

*Figure2: INCEpTION annotation tools.*

Figure 2, shows how spans are marked and source spans and target spans are labelled as defined rhetorical relations and how this is visible on the INCEpTION tool.

### 3.5 Automatic Approches

Hu and Wan (2023) discussed the automatic approaches for discourse parsing with a special focus on the identification and segmentation of discourse units, while their work does not concern the prediction of rhetorical relations among the discourse units. On the contrary, our dataset's created discourse units along with corresponding RST relations and these RST relations predict manually by human annotator, thus enabling us to compare the outcome of human annotation with automatic approaches. In this discussion, we classify human data as gold standard data, that is we are going to train our model on the human-annotated data and compare its performances with the automatic approaches by different machine learning technique.

The comparison will be done using some of the Large Language Models (LLMs), particularly BERT, by Devlin et al. (2019) and Distil-BERT, by Debut et al. (2019), which we will have a look how it is perform in predicting RST relations from the discourse units. Moreover, given that our RST relations are stored in our dataset's as integer labels, we also test a Logistic Regression model for the sake of comparison. The comparative result of these automatic methods evaluated against human-annotated relations would provide insight into the effectiveness of these models in terms of predicting RST relations form the discourse units.

### 3.5.1 Balancing the Dataset

In order to maintain the effective and fair performance of machine learning models for different rhetorical relations, a step that cannot be ignored is the balancing of the dataset. Our study involves the dataset consisting of sports reports which are divided into Elementary Discourse Units (EDUs) annotated with rhetorical relations such as Elaboration, Background, Contrast, Narration, Concession, Restatement, Cause-Effect, and Joint. As it happens with many other NLP research projects, the distribution of these relations does not follow a uniform pattern. Some relations appear more frequently than others, which raises the issue of an imbalanced dataset. A clear display of our dataset imbalance is shown in my Figure 1. An imbalanced dataset can bring about bias in ML models, leading them to favor majority classes while underperforming on minority classes. In our case, relations like Elaboration and Background occur relatively frequently, whereas Concession and Joint are rather rare in the datasets. This imbalance contributes to poor generalization, meaning that the model does have difficulty in correctly predicting less common rhetorical relations. For balancing our datasets, several solutions can be followed: data augmentation, oversampling the minority class, or undersampling the majority class or by using a weighted loss function. However, as our dataset is small, we decided not to undersample the majority class, hence, we have used the oversampling technique. After oversampling our datasets has 25 labels for each rhetorical relations.

### 3.5.2 Logistic Regression

Before delving into the transformer-based model, we will first test our dataset using a logistic regression model, as it is more simpler model. In classification tasks, logistic regression does not accept raw text instead, it requires numerical features derived from the text to help the model identify patterns. The model is trained using labeled data, which is organized into a feature matrix. Below is an example of how a logistic regression-based model learns from the data.

**Example**:-

**EDU1**: "England pace bowler Jofra Archer could play in this year's T20 World Cup but"

**EDU2**: "will not play test cricket until 2025"

**Label**: Contrast

Using numerical feature representations, the logistic regression model is trained to recognize patterns and predict that these EDUs exhibit a Contrast relation.

### 3.5.3 BERT

BERT (Bidirectional Encoder Representations from Transformers) was introduced by Devlin et al. (2019) as a large language model based on the Transformer architecture. It is designed to interpret contextual relationships between words by taking into account both left and right contexts. One of the major tasks to carry out in our project is to evaluate BERT's understanding of rhetorical relations. Employing dynamic embeddings that change according to the wider context of the sentence, this bidirectionality makes BERT particularly attractive for all sorts of natural language processing tasks. We have used the BERT model in our projects for text classification, training with discourse units and their specified relationships and then testing with a test set.

Take for example the following two discourse units:

**DU1**: Archer's England career has been plagued by injuries.

**DU2**: and he has not played a test match since the tour of India in 2021.

A fine-tuned BERT model would classify this pair as "**Elaboration**", since DU2 elaborates on what DU1 says. When comparing the predictions made by BERT and by humans, we can assess the performance of the model and examine areas where it succeeds or fails in capturing rhetorical relations.

### 3.5.4 DistilBERT

DistilBERT (Distilled Bidirectional Encoder Representations from Transformers) is a smaller, faster, and more efficient variant of BERT introduced by Sanh et al. (2019). It is trained using knowledge distillation, a process where a smaller model learns from a larger pre-trained model while retaining most of its performance. DistilBERT maintains higher accuracy while being faster and requiring fewer parameters, making it highly suitable for computationally demanding NLP tasks. For our project, we want to extend our research into DistilBERT model so that we can have a comparision with BERT & DistilBERT model Since DistilBERT retains the core contextual understanding of BERT while being lightweight, it allows for faster training and inference, making it a compelling choice when computational efficiency is a concern. We fine-tuned DistilBERT for our text classification task by training it on manually annotated discourse units and their corresponding relations and then test with the test dataset.

### 3.5.5 Model Training

To train our models, we have two approaches: Logistic Regression and a Transformer-based model. Below, we detail the training process for both models.

**Logistic Regression**: Our datasets contains tokens which is actually encoded datasets and contains inputs ids, attention masks and labels but it cannot be directly applied in logistic regression classifier. Therefore, we need to extract the features, here we have used Transformer model to extract the features. In this method, we freeze its pre-trained weights as we utilize its hidden states as feature representation and extract the last hidden states using PyTorch tensors, effectively capturing the embeddings of the discourse units. The tokenized and processed PyTorch tensors hence give a 768-dimensional vector for each of the discourse units in our dataset's. We build our feature matrix using **scikit-learn** after obtaining the hidden states and subsequently train the logistic regression model with **max iter** set to **3000** to make sure it converges.

**BERT & DistilBERT**: For our Transformer-based model, we train our model using Trainer API & Training arguments from the transformers library. Here, we have defined Training arguments with the key configurations such as:

- **batch size**: 64
- **Number of epochs**: 20
- **Learning rate**: 2e-5
- **Weight decay**: 0.01 for regularization
- **Evaluation strategy**: Performed at the end of each epoch
- **Logging steps**: Dynamically calculated based on dataset size

These are some of the key settings of our model which were used to train both models using the Trainer API.

For the **BERT** model, the training began with a training loss of 2.0100, a validation loss of 2.0573, an accuracy of 0.2250, and an F1

score of 0.1578. As training progressed, the model showed improvements in performance. By the 20th epoch, the training loss decreased to 1.1371, the validation loss reached 1.2201, and the accuracy improved to 0.8500, with an F1 score of 0.8375.

For the **DistilBERT** model, the training started with a training loss of 2.0696, a validation loss of 2.0392, an accuracy of 0.2750, and an F1 score of 0.1406. As training progressed, the model demonstrated improvements in performance. By the 20th epoch, the training loss decreased to 1.3563, the validation loss reached 1.4624, and the accuracy improved to 0.8500, with an F1 score of 0.8464.

### 3.5.6 Model Evaluation

After completing the training phase, the next step is to evaluate the model. To do this, we need to assess it using unseen data. As previously discussed, we have split our datasets into three parts: training, testing, and validation. Now, we will use the validation set to evaluate our model. For our evaluation strategy, we will follow industry-recommended techniques that include accuracy score, F1 score, test loss, and confusion matrix. To generate the evaluation scores for the various metrics, we used the metrics provided by scikit-learn library and for Transformer based model evaluation was conducted using Trainer evaluate function. The evaluate function provided accuracy and F1 scores, where accuracy measures the percentage of correctly predicted labels and the F1 accounts for class imbalances by computing a weighted harmonic mean of precision and recall.

For BERT, the evaluation results were as follows, the loss was 1.2201, the accuracy reached 85%, and the F1 score was 0.8375. For DistilBERT, the loss was 1.4624, the accuracy also the same as BERT and the F1 score was 0.8464.

**Confusion Matrix**

By Calculating the confusion matrix we can get more in details information into our model evaluation.

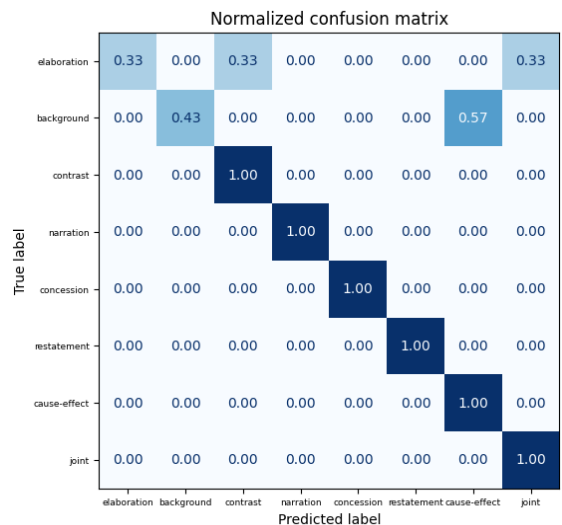

*Figure 3: Confusion Matrix of BERT model*

In Figure 3, we observe that our BERT-based model frequently confuses elaboration relations with contrast and joint relations. Additionally, background relations are often misclassified as cause-effect. This indicates where our BERT model struggles to generalize the relationships effectively.

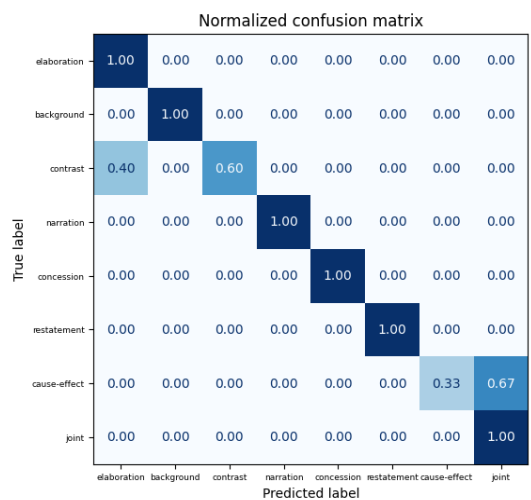

*Figure 4: Confusion Matrix of DistilBERT model*

In Figure 4, we notice that our DistilBERT model often misclassified cause-effect relations as joint relations. Additionally, contrast relations are mostly confused by elaboration relations. This indicates where our DistilBERT model struggles to generalize the relationships.

Our findings indicate that BERT and DistilBERT have the same evaluation accuracy, however In terms of f1 score DistilBERT has a slightly better score. Regarding induvidual relationships, BERT mostly confused on Elaboration and Background relationships, while DistilBERT has difficulty predicting Contrast and Cause-Effect relationships.

## 4. Result

There were two different approaches followed to make judgments about our models, one is the basic evaluation technique, and the other one is the error analysis. For the sake of clarity, we used the test datasets to make the model judgment and the final evaluation.

| Model Name | Accuracy | F1 Score | Test Loss |
|---|---|---|---|
| BERT | 80% | 0.78 | 1.31 |
| DistilBERT | 90% | 0.88 | 1.48 |
| LR (BERT tokenizer) | 97% | - | - |
| LR (DistilBERT tokenizer) | 90% | - | - |

*Tabel 2: Overview of model performance*

In Table 2, we present the evaluation metrics for all models, including accuracy and F1 score. The Transformer-based model, DistilBERT has achieved 90% accuracy and 80% for the BERT model. In contrast, the base model Logistic Regression achieves 97% accuracy when utilizing the BERT tokenizer and 90% when using the DistilBERT tokenizer. F1 score and test loss metrics are available for the Transformer-based models. Specifically, the F1 score for DistilBERT is 0.88, while it is 0.78 for the BERT based model. Based on these evaluation metrics, we can conclude that DistilBERT is the top-performing model for classifying RST relations compared to BERT models. However, its performance improves when we utilize the BERT tokenizer in base models such as logistic regression.

### 4.1 Error Analysis

To gain a better understanding of the model performance, we conducted error analysis by identifying the discourse relations with the highest prediction loss in both BERT and DistilBERT models. This analysis helps pinpoint the most challenging relations for automatic classification and highlights areas for potential improvement.

The top five misclassified relations in the BERT model were examined. The Elaboration relation was misclassified as Contrast, leading to the highest individual loss of 3.2068. Additionally, the Cause-Effect relation was often confused with Narration, suggesting that BERT struggled to distinguish between sequences of events and causal relationships. This indicates that while BERT effectively captures semantic nuances, it may require further fine-tuning to better differentiate relations with overlapping contextual dependencies.

In DistilBERT model, which reveals that the Cause-Effect relation had the highest loss and was frequently confused with Joint relation. This suggests that DistilBERT, being a compressed model, may lack some of the deeper contextual understanding that BERT provides, leading to difficulty in distinguishing between causality and coordination.

## 5. Conclusions

This research focused on annotating rhetorical relations in sports reports using INCEpTION tools and evaluating the effectiveness of automated approaches, such as BERT and DistilBERT. The study showed that manual annotation still plays a vital role in the accuracy and consistency standards of discourse parsing whereas large language models hold potential in automating this task. Results indicated DistilBERT was yielding slightly higher accuracy than BERT, thereby implying that such a small-scale efficient model could perform well on rhetorical relation classification. Additionally, When it came to BERT-based embeddings, the performance of Logistic Regression was quite satisfactory, reaffirming the relevance of classical models in structured datasets. Nevertheless, an error analysis revealed specific difficulties, such as confusion between Elaboration and Contrast, together with Cause-Effect and Narration, which would help to highlight possible areas for further improvement for machine annotations. This work, therefore, is a contribution to discourse parsing studies that connect manual and automated approaches..

While this study provides valuable insights, several areas require further exploration. One potential direction is to add more rhetorical relations and experiment with the vast number of discourse units. Moreover, exploring additional models like GPT, T5, or RoBERTa could also significantly contribute to gaining insights into discourse relation classification. Future works on these areas will help refine automatic approches to be more adaptable, accurate, and useful in natural language processing tasks.

## Acknowledgments

This research was conducted as part of an academic requirement under the supervision of Prof. Dr. Ralf Klabunde, Linguistic Data Science Lab, Ruhr University Bochum.## References

1. Mann, William & Thompson, Sandra. (1988). Rethorical Structure Theory: Toward a functional theory of text organization. University of Southern California.
2. Mann, William & Thompson. (1987b). Rhetorical structure theory: A theory of text organization. Marina del Rey, CA: Information Sciences Institute.
3. Jasinskaja, Katja & Karagjosova, Elena. (2015). Rhetorical Relations. Faculty of Philosophy, University of Cologne.
4. Stede, Manfred & Taboada, Maite & Das , Debopam. (2017). Annotation Guidelines for Rhetorical Structure. University of Potsdam.
5. Taboada, Maite & Mann, William. (2006). Applications of Rhetorical Structure Theory. Simon Fraser University
6. Miltsakaki, Eleni & Prasad, Rashmi & Joshi, Aravind & Webber, Bonnie. (2004). The Penn Discourse Treebank.
7. Klie, Jan-Christoph & Bugert, Michael & Boullosa, Beto & Eckart de Castilho, Richard & Gurevych, Iryna. (2018). The INCEpTION Platform: Machine-Assisted and Knowledge-Oriented Interactive Annotation.
8. Hu, Xinyu & Wan, Xiaojun. (2023). RST Discourse Parsing as Text-to-Text Generation. Peking University.
9. Jacob Devlin, Ming-Wei Chang, Kenton Lee, and Kristina Toutanova. (2019). BERT: Pre-training of Deep Bidirectional Transformers for Language Understanding.
10. Sanh, V., Debut, L., Chaumond, J., & Wolf, T. (2019). DistilBERT, a distilled version of BERT: Smaller, faster, cheaper and lighter.
11. Asher, Nicholas., & Lascarides, Alex. (2003). Logics of conversation. Cambridge University Press.